\documentclass[11pt,a4paper]{article}
\usepackage[hyperref]{emnlp2020}
\usepackage{times}
\usepackage{latexsym}

\usepackage{microtype}

\aclfinalcopy 

\usepackage{booktabs}
\usepackage{graphicx}

\usepackage[symbol]{footmisc}

\title{A New Neural Search and Insights Platform for\\ Navigating and Organizing AI Research}

\author{Marzieh Fadaee\footnote[1]{} \quad Olga Gureenkova\footnote[1]{} \quad Fernando Rejon Barrera\footnote[1]{} \\
\textbf{Carsten Schnober\footnote[1]{} \quad Wouter Weerkamp\footnote[1]{} \quad Jakub Zavrel}\thanks{\enspace All authors contributed equally.}  \\
Zeta Alpha\\
\texttt{\{lastname\}@zeta-alpha.com}}

\date{}

\begin{document}
\maketitle

\begin{abstract}
To provide AI researchers with modern tools for dealing with the explosive growth of the research literature in their field, we introduce a new platform, AI Research Navigator, that combines classical keyword search with neural retrieval to discover and organize relevant literature. The system provides search at multiple levels of textual granularity, from sentences to aggregations across documents, both in natural language and through navigation in a domain specific Knowledge Graph. We give an overview of the overall architecture of the system and of the components for document analysis, question answering, search, analytics, expert search, and recommendations.
\end{abstract}

\section{Introduction}

\begin{figure}[t!]
    \centering
    \includegraphics[width=1.0\columnwidth]{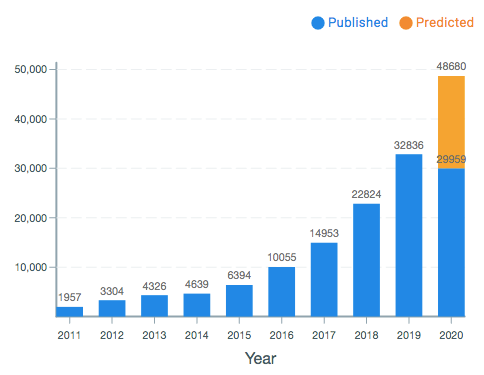}
    \caption{Growth of AI related documents on arXiv.}
    \label{fig:arxivgrowth}
\end{figure}

The growth of publications in AI has been explosive in recent years. 
A big portion of this growth is happening on platforms outside of traditional publishing venues, for instance arXiv e-print archive (see Figure~\ref{fig:arxivgrowth}) and blogs. Although this encourages broad access to AI expertise and technology, it makes efficient and effective search, monitoring, and discovery in the AI field increasingly difficult. Most general-purpose academic search engines lack a specialization on AI content and practical know-how, because they focus on classical bibliographic information across all scientific disciplines. 
At the same time, academic search engines often do not make use of the latest AI technologies in search, as well as natural language processing (NLP) and insights capabilities.
The main reason that limits them is the need to operate at a much larger scale and cover a large amount of knowledge.

Recent developments in various NLP tasks are showing fast progress towards an almost human level of language understanding \cite{devlin-etal-2019-bert,brown2020language}. Applying these new technologies to the processing of research and engineering literature bears the promise of accelerating scientific discovery.
In addition, providing efficient tools to automate some of the drudgery of human scholarly work by machine understanding of scientific knowledge is extremely valuable. Similar directions are being explored in recent studies \cite{ammar2018construction,axcell,zhao-lee-2020-talk}.

Our system, AI Research Navigator\footnote[2]{\texttt{search.zeta-alpha.com}}, aims to help AI researchers with a simple-to-use semantic search for documents (\S\ref{sec:search}), the answering of detailed factual questions (\S\ref{sec:qa}), the generation of insights via visual analytics (\S\ref{sec:analytics}), combined with recommendations to filter the constant flood of new information in their field (\S\ref{sec:recsys}). These technologies are combined in the platform with both project and task-oriented tools to support a more effective and efficient organization of a researcher's work on multiple projects and topics (\S\ref{sec:organization}). This paper presents a short outline of the system.


\section{Document Analysis}
\label{sec:processing}

A key ingredient to an AI insights platform is the content available to users. 
We ingest, process, and store documents from a variety of sources, aiming to get broad coverage, as well as detailed views on theoretical and applied AI. 
At the moment our platform contains approximately 140 thousand scientific papers, collected from \texttt{arXiv.org}~\footnote{We include all papers from the following AI related categories: cs.AI, cs.LG, cs.CV, cs.IR, CS.NE, cs.CL, and stat.ML.} and \texttt{OpenReview.net}, and about 24 thousand posts from data science blogs. Our goal for the near future is to expand this set of sources to include different types of content (e.g., source code, news, tweets).
%
We ingest new content from our sources on a daily basis, offering our users the latest insights. 
Newly ingested or updated content is fed to our back-end storage system (Section~\ref{sec:storage}) via a distributed processing pipeline that takes the documents through a number of processing and information extraction steps and generates embeddings that capture the intent and meaning of the text. We also extract images from each document to serve as an illustration for the paper in the search engine. We manage the state of each document, as it traverses the pipeline, with a messaging queue platform (Apache Pulsar). 
This allows us to scale our processing throughput, while keeping certain processing guarantees. 

\subsection{Parsing and Linking Documents}

Scientific publications consist of sections that have varying degrees of informativeness. As an example, the bibliography of a paper is interesting for the citation graph, but does not contain actual new content. 
In order to process and index only relevant and informative sections, we parse the document structure and extract candidate citation records using ParsCit~\cite{councill-etal-2008-parscit}. We then sanitize these candidates using a set of heuristics, and link them to our Knowledge Graph (KG, Section~\ref{sec:storage}) using fast approximate string matching. 

%

\begin{figure}[htb]
	\centering
	\includegraphics[width=0.45\textwidth]{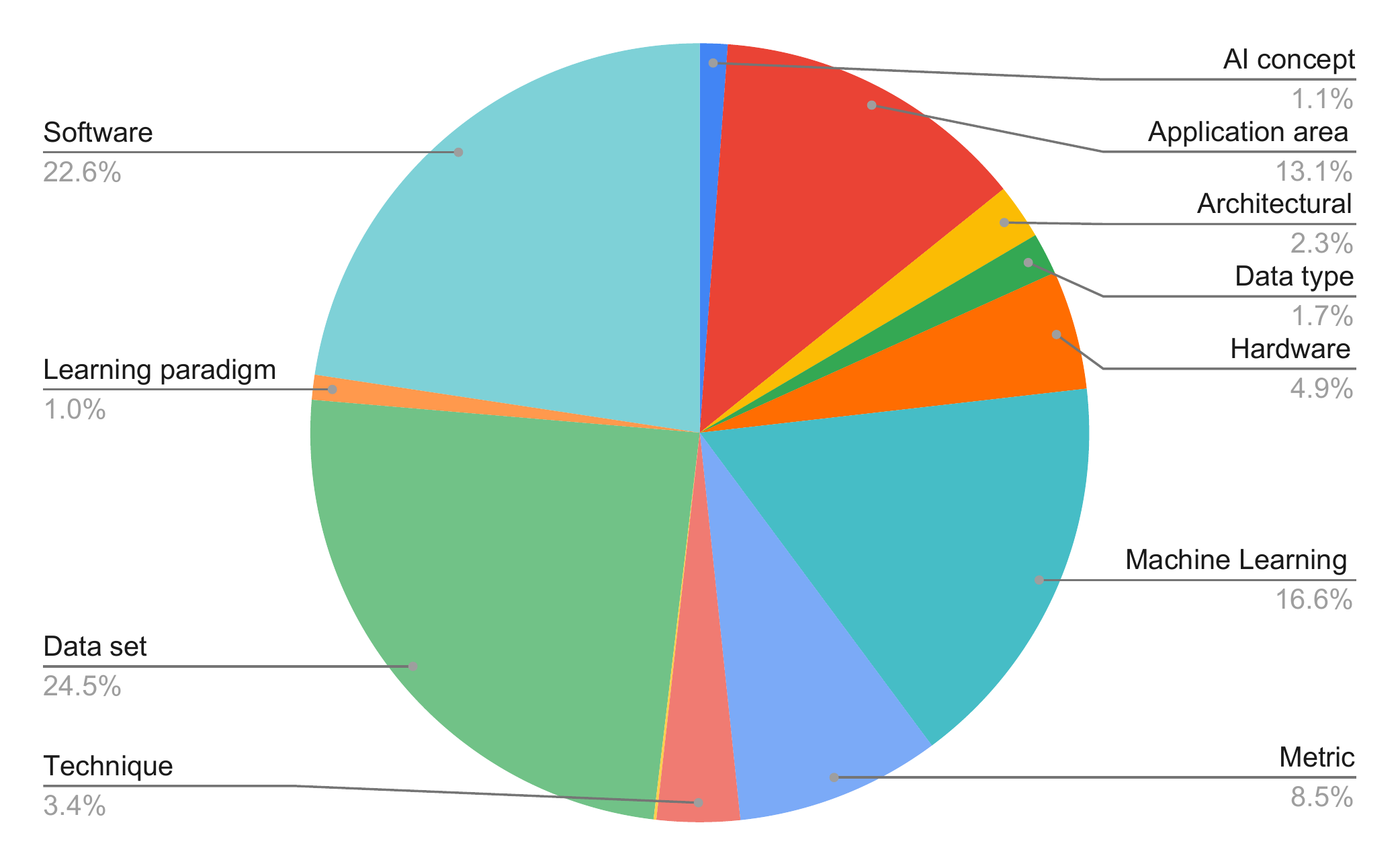}
	\caption{Distribution of KG concept types.}
	\label{fig:kgstats}
\end{figure}

We make use of domain-specific concepts and their relations to improve the effectiveness of components like Question Answering (QA, Section~\ref{sec:qa}), KG population, analytics processing (Section~\ref{sec:analytics}), and semantic search (Section~\ref{sec:search}). 
We train a statistical named entity recognizer (NER) using a small manually curated seed set of around a thousand AI related concepts, and run this NER on every document that we process. 
We then link the recognized concepts to concepts in our KG using a weighted combination of string and contextual embedding similarity. 
Finally, those entities that were not linked because they fell below a similarity threshold are considered as new candidates to further populate the KG. 
Figure~\ref{fig:kgstats} shows the domain-specific concept types that are currently in our KG.

\section{Storage Systems}
\label{sec:storage}
To provide access to the information we obtained from our processing pipeline, we currently store our information in three core systems.

\paragraph{Knowledge Graph.} We store people, content, and concepts in a knowledge graph (stored in Dgraph). The node identifiers are used in the search frontend for navigating content and building queries.

\paragraph{Search Index.} For fast access to documents, we use an open-source search engine (Elasticsearch) in combination with HNSW~\cite{DBLP:journals/corr/MalkovY16} for nearest neighbor search. We use three separate indices, one for sentences (31M), one for chunks (approx. 10 consecutive sentences) (4.2M), and one for full documents (160K) and citation records (740K).

\paragraph{Document Representations.} As a result of each processing step, we store new document representations. Most representations are in the form of standoff annotations, linking labels (e.g., a concept ID or vector) to a particular span of characters in the source document.

\section{Accessing Information}
\label{sec:access}
Processing and storing information is only useful when we can provide meaningful access to it. 
Our platform allows information access in a variety of ways. 
In this section, we discuss content (\S\ref{sec:search}) and expert (\S\ref{sec:expert}) search, QA system (\S\ref{sec:qa}), analytics component (\S\ref{sec:analytics}), and recommendations system (\S\ref{sec:recsys}).

\subsection{Content Search}
\label{sec:search}

One of the main methods to access information on our platform is search. We currently support traditional keyword-based search and vector-based (nearest neighbor) search. 
Both search systems offer valuable information.
While keyword-based search is useful in finding documents directly related to the query, vector-based search offers a range of query interpretations and more diversity. 

\paragraph{Keyword-based search.}
Our keyword-based search functionality scores documents for a given user query based on several heuristics. 
(1)~We borrow from \citet{10.1145/1076034.1076115} the notion of sequential dependencies between query terms, construct term $n$-grams from the user query, and treat each $n$-gram as a phrase query. 
(2)~A match of a longer $n$-gram is more important than the match of a shorter $n$-gram, which is implemented as a dynamic boost per $n$-gram query. 
(3)~We combine evidence from multiple document fields~\cite{10.1145/860435.860463} and assign higher weights to metadata fields like author name, title, and abstract, while limiting the weight of the full text field. 
(4)~We use a \texttt{dismax} query over fields to determine whether an $n$-gram refers to an author or to content. 
(5)~Given the limited text length of the metadata fields, we only rely on term presence, and assign a constant field-dependent score. 
(6)~Finally, we assume that highly cited and recent documents are more important to users.

\paragraph{Vector-based.}
In many cases, keyword searches are hard to use when exploring a new domain. 
To allow a more meaning-based exploration, fully neural retrieval models can be beneficial. Recent advances in neural language modeling as unsupervised pretraining have achieved significant improvements in a wide variety of NLP tasks \cite{devlin-etal-2019-bert}.
However, incorporating them in retrieval systems presents some challenges.
%
Using pretrained language models to jointly encode queries and documents is often not computationally feasible for large-scale retrieval. 
Recent studies propose various methods to benefit from large neural models. 
\citet{luan2020sparse} propose a hybrid method for combining sparse and dense representations that outperforms baselines in open retrieval. \citet{Chang2020Pre-training} use a siamese network, initialized with BERT, to encode query and document individually. 
They propose three self-supervised tasks that capture different aspects of query-document relations.

Inspired by these studies, we use the Sentence-BERT model proposed by \citet{reimers-gurevych-2019-sentence} to generate sentence embeddings for each document and, additionally, we also encode all words in context with SciBERT \cite{Beltagy2019SciBERT} embeddings. Finally, we encode sentences, chunks, and full documents into representative vectors, fine-tuned on self-supervised training tasks similar to~\citet{Chang2020Pre-training}.
In our platform, we encode the query as a vector at search time, and retrieve its $N$ (approximate) nearest neighbor documents in the vector space. This  requires the documents and the queries to be encoded in a similar way using the same embedding space. By using HSNW and loading its full graph in memory, we are able to serve nearest neighbor search results in a highly efficient manner. 

\subsection{Expert Search}
\label{sec:expert}

In addition to navigating knowledge via natural language search and domain-specific topics from our KG, we also aim to improve navigation by connecting searchers to experts. For this, we follow a document-centric approach to expertise retrieval, along the lines sketched in~\citet{balog2009} and~\citet{husain2019}. 
In the expert search component, we embed user queries and documents using the Sentence-BERT model similar to Section~\ref{sec:search}. 
This allows the system to retrieve papers that are related to the query. 
We then derive the experts from the sets of authors of these papers using an approach where each retrieved paper contributes an exponentially weighted vote for an author, with a factor that reduces the bias towards highly prolific authors. 
Our experiments, described in detail in~\cite{markberger2020b} show that these modern Transformer-based contextualized embeddings outperform TF-IDF and LSI-based document representations on this task.

\subsection{Question Answering}
\label{sec:qa}

\begin{figure}[htb!]
    \centering
    \includegraphics[width=.8\columnwidth]{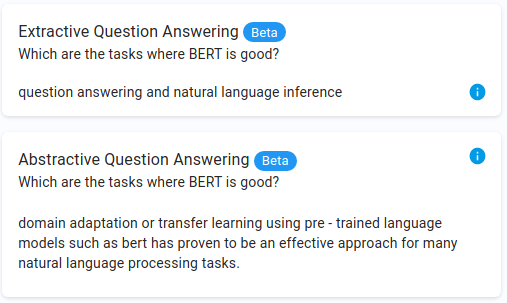}
    \caption{Extractive and abstractive QA components.}
    \label{fig:ex-ab-qa}
\end{figure}

Our QA module provides an answer to either a concrete user question (e.g., \textit{How many TPUs are needed to train BERT?}) or, alternatively, to a question related to the user's query, which we automatically generate (e.g., \textit{What is a knowledge graph?} for the query \textit{knowledge graph}). To distinguish between questions and other types of queries, we use a Naive Bayes classifier trained on the NPS Chat Corpus~\cite{NPS:2006}.



We deploy two types of QA deep learning models: extractive QA and abstractive QA (see example in Figure~\ref{fig:ex-ab-qa}). Both models take as input a set of relevant documents, and provide the user with one or more answers. Since an answer is always part of a particular context, we also present this context as a source of explanation of the answer to the user.


\paragraph{Extractive QA.} Our extractive QA model is built with an existing BERT-based question answering model from the DeepPavlov library~\cite{burtsev-etal-2018-deeppavlov}. 
The model takes as input a pair of question and context and rate their relatedness. 
At query time, we chunk the input documents and send multiple question-context pairs to the model. 
We obtain the best answer by filtering the candidates according to the confidence of the model.

\paragraph{Abstractive QA.} For the abstractive answers we use a model based on the approach proposed in~\citet{nishida2019multistyle}, trained on the MS-MARCO data set~\cite{bajaj2016ms}. Our model and its evaluation in the AI domain are described in detail in~\citet{tsiamas2020}.
Since this architecture has its own neural retrieval component, at query time 
the model has access to the question and all input documents.
Unlike extractive QA, this model is also capable of answering yes/no questions. 

The two QA models complement each other in the types of answers they provide.
Although these systems are still experimental (approximately 70\% of answers to a benchmark set of in-domain questions were relevant), together with sentence and paragraph retrieval they show potential for discovering interesting information that goes beyond what surface-level single-document-based systems provide.





\subsection{Analytics}
\label{sec:analytics}

Rather than reviewing a long list of documents or reading a short answer in response to a query, sometimes users can get to an insight faster by observing a tabular, a summary, or a graph overview 
over the entire set of relevant documents. 

\begin{figure}[htb!]
    \centering
    \includegraphics[width=0.8\columnwidth]{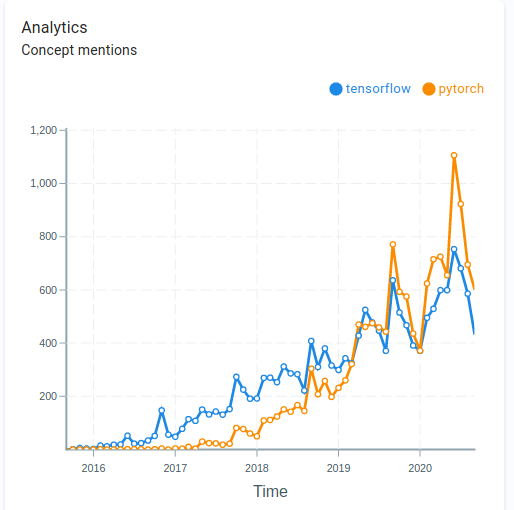}
    \caption{Contrastive popularity analytics.}
    \label{fig:plot-contrastive}
\end{figure}

Our analytics module aims to give users this quick and easy-to-grasp overview over a result set when a query sufficiently matches some pre-defined analytics query templates. For instance, when we detect AI concepts 
from the KG in a query, or a reference to an abstract concept (e.g.,~{\it ``Which datasets are used for image classification?"}), we show the contrastive popularity plots for the specific concepts as identified in documents using the NER and linker module. 
Figure~\ref{fig:plot-contrastive} provides an example of a contrastive popularity graph.
The graphs provide a global overview and are also clickable so the users can use them to quickly identify patterns and discover specific papers relevant to their interests. 

\subsection{Recommendations}
\label{sec:recsys}

\begin{figure}[htb!]
    \centering
    \includegraphics[width=.8\columnwidth]{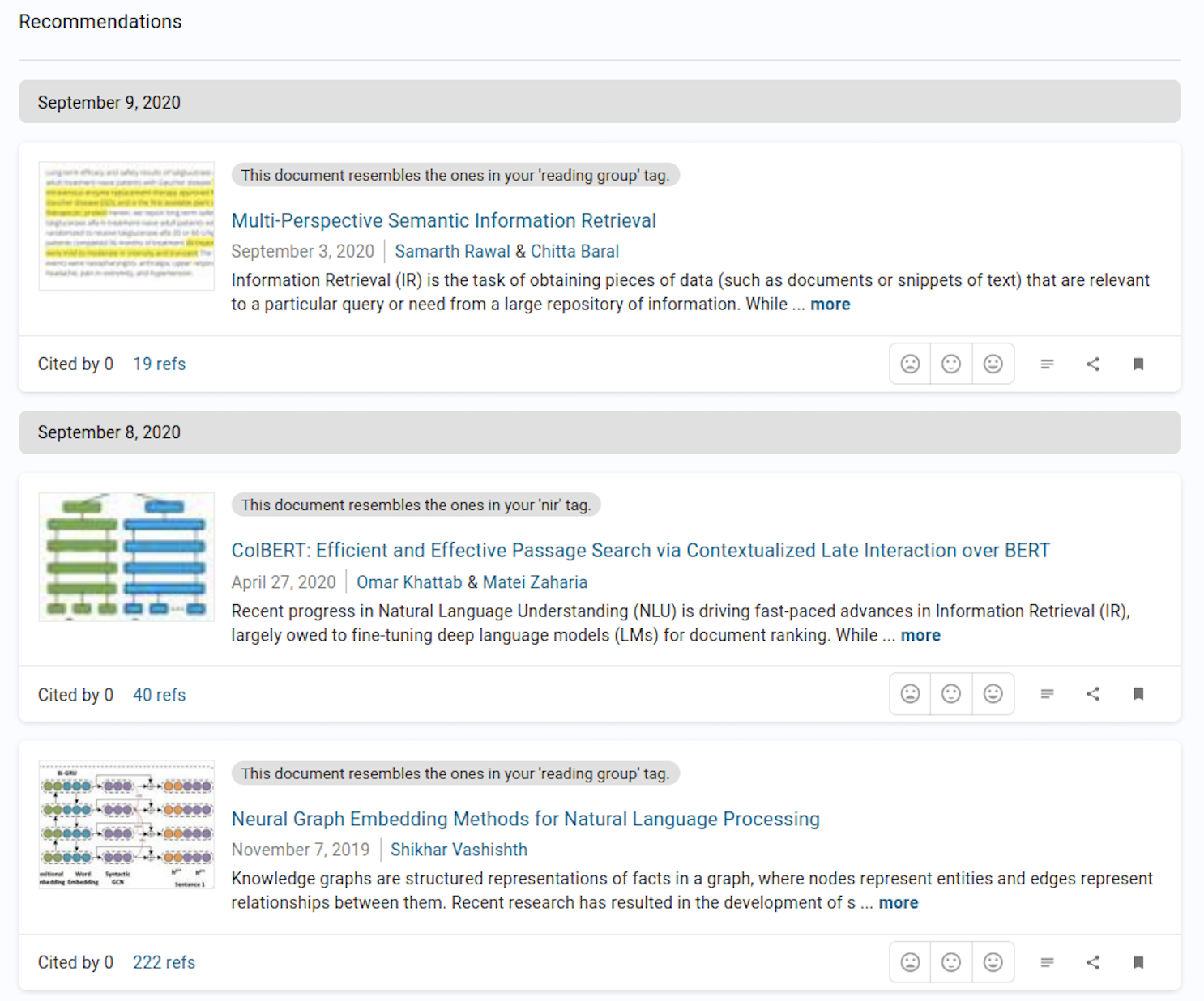}
    \caption{Recommendations on our platform.}
    \label{fig:recommendations}
\end{figure}

With the amount of new information available on a daily basis, a recommender system is inevitable to filter and keep track of relevant publications. Users of our platform receive recommendations in notification emails and in the recommendations view (Figure~\ref{fig:recommendations}).

The relevance of a publication can be decomposed into several factors. 
We implemented a modular system architecture which allows us to weigh relevance factors on a per-user basis, and to add, remove, modify, and evaluate modules individually. Each module generates recommendations with (normalized) scores, which are aggregated by the core recommender system, allowing for both global and personalized weights for each module.

Our recommendation architecture is based on hybrid recommender systems~\cite{gomez-uribe_netflix_2016}, combining content-based and collaborative filtering. 
However, since virtually all recommendations are of new papers, we suffer from the cold-start problem and we mostly rely on content-based recommendations.
The content-based module generates recommendations based on user-tagged documents: 
when a user tags a document, it triggers an initial search for related documents. 
From this point on, recommendations are only generated from the most recent documents.

Our current basic content-based modules are based on similarity metrics derived from our document representations, as described in section~\ref{sec:search}, with score normalization being provided by leave-one-out tuning on the set of documents in a tag.

Apart from these content-based similarity recommendations, we are also experimenting with additional modules that provide similarity scores. \textbf{Citation-based} recommendations are based on (indirect) citations to documents stored by the user. \textbf{Author-based} recommendations are (co-)authored by authors which are frequently tagged by a user. \textbf{Popularity} recommendations are globally ``popular'' documents, for instance based on the number of views, citation counts, or tag counts.

\section{Productivity Tools}
\label{sec:organization}
\begin{figure}
    \centering
    \includegraphics[width=\columnwidth]{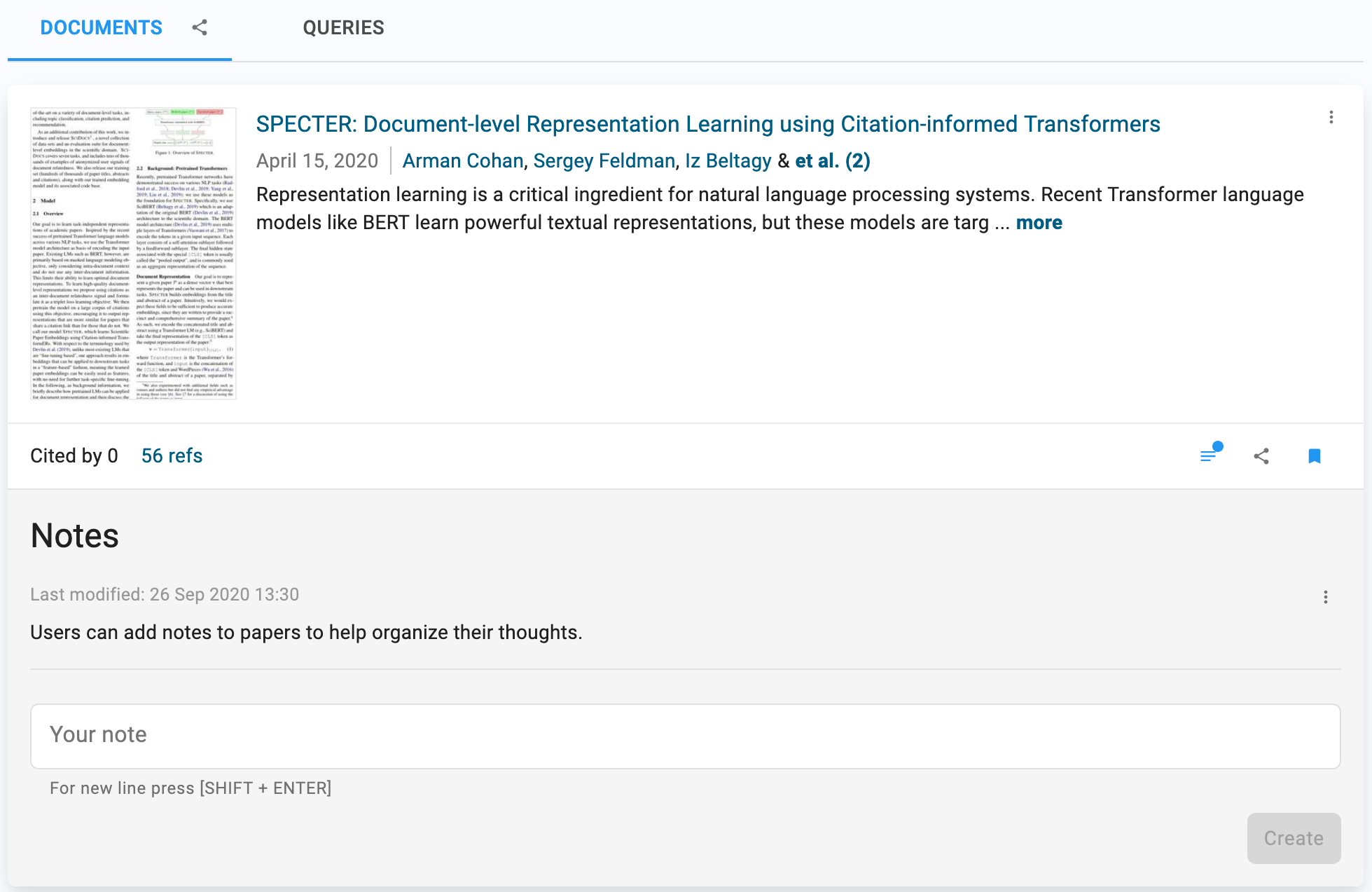}
    \caption{Adding tags and notes to a document.}
    \label{fig:notes}
\end{figure}

Discovering relevant information in an effective way is key to researchers, knowledge workers, and decision makers. 
Even though an AI-enabled platform like the one described in this paper can be helpful for this purpose, it is only the first step in researching a topic.
Organizing and accessing this information is a necessary feature. 
Users of our platform are also supported to organize information and knowledge without having to rely on external tools for reading lists and notes. 
Having found a relevant piece of information in our system, 
users can save this into their own specific project and topic tags. They can also directly write their notes on the papers and projects they are working on in the tool.
Tags serve to organize lists of documents, as well as the queries used, and notes taken while working on a project 
(see Figure~\ref{fig:notes}). The tagging system can also be used to track the status and priority of work. Tag-based lists can easily be shared with others within the platform, on social media, and exported into other tools. 
As described above, these tags are also the starting point that allows users to be alerted about new results relevant to their interests. 

\section{Discussion and Next Steps}


Having introduced a new platform to discover and organize knowledge for AI researchers, we foresee considerable future research to reach real machine understanding of scientific literature, such as extraction of complex entity relations and more advanced use of neural embeddings to reduce the dependency on manual KG curation. 
We leverage a mix of state-of-the-art AI components to give researchers transparent access to a body of knowledge from a large volume of heterogeneous and non-reviewed content.
As a result, it raises the concern of dealing with fairness, factuality, conflicting opinions, and out-of-date information, which requires deeper investigation. 
Finally, we are interested to further explore how the productivity tools in our platform can contribute to better collaboration in teams and improving knowledge sharing and discovery.

\section*{Acknowledgements}
We would like to acknowledge the joint effort
from Zeta Alpha's team including Shamil Mammadov and Victor Zuanazzi to make this
work possible.

\bibliography{emnlp2020}
\bibliographystyle{acl_natbib}

\end{document}